\pdfoutput=1

\documentclass[11pt]{article}

\usepackage{emnlp2021}

\usepackage{times}
\usepackage{latexsym}
\usepackage{graphicx}
\usepackage{amsmath}
\usepackage{bbm}
\usepackage{multirow}
\usepackage{comment}
\usepackage{hhline}
\usepackage{booktabs}
\usepackage{enumitem}

\usepackage[T1]{fontenc}

\usepackage[utf8]{inputenc}

\usepackage{microtype}

\newcommand{\aside}[1]{}            

\title{Neural Attention-Aware Hierarchical Topic Model}

\author{Yuan Jin$^{1}$ ~~~~~ He Zhao$^{1}$ ~~~~~ Ming Liu$^{2}$ ~~~~~ Lan Du$^{1}$\thanks{Corresponding author} ~~~~~ Wray Buntine$^{1}$\\
$^{1}$Faculty of Information Technology, Monash University, Australia \\
{\tt \{yuan.jin, ethan.zhao, lan.du, wray.buntine\}@monash.edu} \\
$^{2}$School of Information Technology, Deakin University, Australia \\
{\tt m.liu@deakin.edu.au} \\}

\begin{document}
\maketitle
\begin{abstract}
\textit{Neural topic models} (NTMs) apply deep neural networks to topic modelling. Despite their success, NTMs generally ignore two important aspects: (1) only document-level word count information is utilized for the training, while more fine-grained sentence-level information is ignored, and (2) external semantic knowledge regarding documents, sentences and words are not exploited for the training. To address these issues, we propose a variational autoencoder (VAE) NTM model that jointly reconstructs the sentence and document word counts using combinations of bag-of-words (BoW) topical embeddings and pre-trained semantic embeddings. The pre-trained embeddings are first transformed into a common latent topical space to align their semantics with the BoW embeddings. Our model also features hierarchical KL divergence to leverage  embeddings of each document to regularize those of their sentences, thereby paying more attention to semantically relevant sentences. Both quantitative and qualitative experiments have shown the efficacy of our model in 1) lowering the reconstruction errors at both the sentence and document levels, and 2) discovering more coherent topics from real-world datasets.

\end{abstract}

\setlength{\abovedisplayskip}{3pt}
\setlength{\belowdisplayskip}{3pt}

\section{Introduction}
Topic models are a family of powerful techniques that can effectively discover human-interpretable topics from unstructured corpora for text analysis purposes. Among them, Bayesian topic models, based on the latent Dirichlet allocation (LDA) \cite{blei2003lda}, have been the mainstream for nearly two decades. They usually adopt count/bag-of-words (BoW) representations for text content and model the generation of the BoW data with a probabilistic structure of latent variables. These variables follow pre-specified distributions under Bayes’ theorem, and are learned through Bayesian inference such as variational inference (VI) and Monte Carlo Markov chain (MCMC) sampling. Despite their success, conventional Bayesian topic models, however, are known to lack flexibility in their model structures and scalability to large volumes of data. 

To address the above limitations, increasing effort has been made in leveraging deep neural networks (DNNs) for topic modelling, which leads to the so-called neural topic models (NTMs) \cite{zhao2021topic}. Most of these models follow the framework of variational auto-encoders (VAEs)~\cite{kingma2013auto,rezende2014stochastic} and adopt an encoder-decoder architecture, in which the encoder transforms the BoW data of each document into the corresponding document-topical embeddings, and the decoder attempts to map these embeddings back to the same data. 
With a moderate increase in model complexity, NTMs have largely outperformed conventional topic models on BoW data reconstruction and topic interpretability/coherence \cite{pmlr-v48-miao16, Srivastava_Sutton_2017,ding-etal-2018-coherence,zhou-etal-2020-neural,zhao2020neural}.

With this being said, most NTMs only exploit the BoW information of internal documents while ignoring (1) the sentence-level BoW information of these documents, and (2) the external (semantic) information regarding the documents, sentences and words (e.g., extracted from other larger relevant corpora). These limitations have hindered the further performance improvement for NTMs. Therefore, in this paper, we propose a new NTM that address these limitations. It jointly reconstructs the BoW data of each document and their sentences with combinations of both internal BoW topical embeddings and external pre-trained semantic embeddings. To do this, we design an \textit{internal BoW data encoder} and an \textit{external knowledge encoder} to respectively transform the BoW data and the pre-trained embeddings of the same documents and sentences into a shared latent topical space. The resulting internal and external topical embeddings are then combined to decode the BoW data.

To address the BoW data sparsity \cite{zhao2019leveraging} at the sentence level, which has been a problem for many topic models, our model enforces hierarchical KL divergence on pairs of sentences and corresponding documents with respect to their topical embeddings derived from both the BoW data and the external knowledge. The intent is that both types of topical embeddings for each sentence should be governed by the same types of embeddings of their parent documents. 
Furthermore, the hierarchical KL for a sentence is weighted by its semantic relations to its parent document, so that a document's topical information is more influential to more semantically representative sentences.
Our contribution can be summarized as follows: 
\begin{itemize}[leftmargin=*,noitemsep, nosep]
    \item We propose a VAE-based neural topic model, which encodes internal BoW information and external semantic knowledge specific to word, sentence and document levels into the same latent topical space to refine topic quality.
    \item Our model imposes attention-weighted hierarchical KL divergence on pairs of sentences and documents to smooth the learning of the topical embeddings from sparse BoW data of sentences.
    \item We demonstrate that our model is effective in BoW data reconstruction at both the document and sentence levels. 
    It also improves the internal and external coherence of the discovered topics.
\end{itemize}

\section{Related Work}\label{sec:related_work}
To our best knowledge, most state-of-the-art NTMs, like ours, are based on the VAE framework. For a detailed discussion of these NTMs, we refer readers to \citet{zhao2021topic}. Here, we only discuss the two lines of research that are most relevant to ours.

\paragraph{NTMs with pre-trained language models}
Recently, pre-trained transformer-based language models such as BERT~\cite{devlin2018bert} have started to draw the attention of the topic modelling community thanks to their capability of generating contextualized word embeddings with rich semantic information that is absent from the BoW data. Thus, an emerging trend focuses on incorporating these contextualized word embeddings into NTMs. 

Based on the popular VAE framework of \citet{Srivastava_Sutton_2017},
\citet{bianchi-etal-2021-pre} proposed a contextualized topic model (CTM) that incorporates pre-trained document embeddings generated by sentence-transformers~\cite{reimers2019sentence}. However, unlike ours, CTM ignores the other levels of pre-trained knowledge and BoW information. \citet{chaudhary2020topicbert} proposed to combine an NTM with a fine-tuned BERT model by concatenating the topic distribution and the learned BERT embedding of a document as the features for document classification. \citet{hoyle-etal-2020-improving} proposed BAT, a NTM framework ``taught'' by external knowledge distilled from a pre-trained BERT model. BERT predicts probabilities for each word of a document which are then averaged to generate a pseudo BoW vector for the document. The BAT framework can be instantiated with various existing NTMs such as Scholar~\cite{card-etal-2018-neural} (i.e. BAT+Scholar), which imposes a logistic Normal as the variational posteriors for document embeddings in the VAE, and W-LDA~\cite{nan-etal-2019-topic} (i.e. BAT+W-LDA), which replaces the KL divergence with the maximum mean discrepancy \cite{JMLR:v13:gretton12a}.

\paragraph{NTMs for modelling document structures} Although document structures, i.e., the structured relationships between documents, paragraphs, and sentences have been modelled in conventional Bayesian topic models (e.g., in~\citet{du2012sequential,balikas-etal-2016-sentlda,jiang2019sentence}), they have not been carefully studied in NTMs to our knowledge.
The closest work to our idea is \citet{Nallapati_Melnyk_Kumar_Zhou_2017}, which proposes an NTM that samples a topic for each sentence of an input document and then generates the word sequence of the sentence with an RNN conditioned on the sentence's topic. However, this work focuses on sentence generation instead of topic modelling.

\section{Preliminaries}
In this section, we introduce the VAE framework of neural variational topic models, based on which our model will be developed. 
Table~\ref{table:notation} details the notations and symbols used throughout the paper.

\begin{table}[t]
\setlength\tabcolsep{4pt}
\centering
\small
\begin{tabular}{|p{2cm}|p{5cm}|}
\hline
{\bf Symbols} & {\bf Description}\\
\hline
 $I$, $V$, $K$, $L$, $J_i$, $M$ & size of corpus, vocabulary, topics, hidden neurons, number of sentences in document $i \in (1,...,I)$ and pre-trained embedding dimension\\
 $\boldsymbol{N}$&a vector of the lengths of each document\\
 $\boldsymbol{w}^{\text{D}}_i,\boldsymbol{w}^{\text{S}}_{ij}$ & BoW vectors of the $i^{\text{th}}$ document and its $j^{\text{th}}$ sentence, ($\mathbbm{Z}^{V}_{\geq 0}$)\\  
  \hspace*{-2pt}$\boldsymbol{W}^{\text{D}},\boldsymbol{W}^{\text{S}}_i$ & 
  BoW matrices over the $I$ documents ($\mathbbm{Z}^{I\times V}_{\geq 0}$) and the $J_i$ sentences of document $i$ ($\mathbbm{Z}^{J_i\times V}_{\geq 0}$)\\
  $\boldsymbol{X}^{\text{W}}, \boldsymbol{X}^{\text{S}}_i,\boldsymbol{X}^{\text{D}}$ & matrices of pre-trained non-contextual word embeddings ($\mathbbm{R}^{V\times M}$), pre-trained sentence embeddings ($\mathbbm{R}^{J_i\times M}$) and document embeddings ($\mathbbm{R}^{I\times M}$)\\
  $q^{h}(\cdot)$ & the VAE encoder for $h\in\{\text{``B''}, \text{``E''}\}$ that generate the topical embeddings of words, sentences and documents \\
  $f^{h}(\cdot)$&a shared MLP of the encoder for $h$ that maps $\mathbbm{R}^{V}$ to $\mathbbm{R}^{L}$\\
    $l^{h}_{1}(\cdot), l^{h}_{2}(\cdot)$&linear layers over the shared MLP of the encoder for $h$ that maps  $\mathbbm{R}^{L}$ to $\mathbbm{R}^{K}$\\
  $\boldsymbol{\Phi}^{h}$ & word-topic embedding matrix over each word specific to the encoder for $h,~(\mathbbm{R}^{V\!\times\! K})$\\
    $\boldsymbol{s}^{h}_{ij},\boldsymbol{S}^{h}_i$&sentence-topic embedding of the $j^{\text{th}}$ $(1,...,J_i)$ sentence in document $i$ generated by the encoder for $h,~(\mathbbm{R}^{K})$; sentence-topic matrix, $(\mathbbm{R}^{J_i\times K})$\\
    $\boldsymbol{z}^{h}_i,\boldsymbol{Z}^{h}$ &topical embedding of document $i$ generated by the encoder for $h$, ($\mathbbm{R}^{K}_{\geq 0}$); document-topic matrix, $(\mathbbm{R}^{I\times K}_{\geq 0})$\\
    $\alpha^{\text{S}}_{ij}$&attention weight of sentence $j$ for document $i, (\mathbbm{R})$\\
\hline
\end{tabular}
\caption{List of Notation used. Here, $h$ is either ``B'' or ``E'' representing respectively the internal BoW data and the external semantic knowledge.} 
\label{table:notation}
\end{table}

\paragraph{Problem Formulation}
Consider a corpus that consists of $I$ documents where the $i^{\text{th}}$ ($1,...,I$) document is represented as a $V$-dimensional vector of word counts, $\boldsymbol{w}^{\text{D}}_i$, also known as the bag-of-words (BoW) data. Here, $V$ is the size of the vocabulary from the corpus, and 
$w^{\text{D}}_{iv} $ is the number of times the $v^{\text{th}}$ ($1,...,V$) word occurs in the $i$-th document. Topic modelling assumes that there exist $K$ topics that can be used to describe each document. Its goal is to recover these topics from the BoW data of the corpus $\boldsymbol{W}^{\text{D}}$. In this paper, we further formulate the topic modelling problem at the level of sentences; that is the $i^{\text{th}}$ document comprising a total of $J_i$ sentences. In this case, the document can be alternatively represented as a word count matrix $\boldsymbol{W}^{\text{S}}_i$, an additional level of BoW information we aim to leverage. The $j^{\text{th}}$ ($1,...,J_i$) row of this matrix accommodates the $V$-dimensional BoW vector $\boldsymbol{w}^{\text{S}}_{ij}$ of the $j^{\text{th}}$ sentence from the $i^{\text{th}}$ document.

\paragraph{Neural Variational Topic Model}
Most traditional topic models are graphical models with exponential family model parameters. Therefore, they yield tractable inference for the posterior distributions of the model parameters. However, the inference is limited in expressiveness, thereby less capable of capturing the true underlying generative process and distributions. To solve this problem, neural variational inference is introduced into topic models, named neural variational topic models (NVTM), where more expressive posterior distributions are constructed for the model parameters using neural networks.
A typical NVTM is learned by maximizing the Evidence Lower Bound (ELBO) of the marginal likelihood of the BoW data $\boldsymbol{W}^{\text{D}}$ of each document with respect to their topical embeddings $\boldsymbol{Z}^{\text{B}}$:
\begin{equation}
\begin{split}
 \mathbbm{E}_{\boldsymbol{Z}^{\text{B}}\sim{q}^{\text{B}}(\boldsymbol{W}^{\text{D}})}&\big[\text{log}~{p}(\boldsymbol{W}^{\text{D}}|\boldsymbol{Z}^{\text{B}})\big]\\&-\text{KL}\big[{q}(\boldsymbol{Z}^{\text{B}}|\boldsymbol{W}^{\text{D}})||{p}(\boldsymbol{Z}^{\text{B}})\big]
\end{split}
\end{equation}where $\text{log}~{p}(\boldsymbol{W}^{\text{D}}|\boldsymbol{Z}^{\text{B}})$, ${q}(\boldsymbol{Z}^{\text{B}}|\boldsymbol{W}^{\text{D}})$ and ${p}(\boldsymbol{Z}^{\text{B}})$ are respectively the BoW data likelihood, the variational posterior distribution and the prior distribution of $\boldsymbol{Z}^{\text{B}}$. A key characteristic of NVTM is to model the first two terms as decoder and encoder networks. The encoder network ${q}^{\text{B}}(\boldsymbol{W}^{\text{D}})$ models the posterior distribution ${q}(\boldsymbol{Z}^{\text{B}}|\boldsymbol{W}^{\text{D}})$ by taking in the BoW data $\boldsymbol{W}^{\text{D}}$ to parameterize diagonal 
Gaussian distributions\footnote{In this paper, we follow the original VAE's parametrization of the posterior distribution as a diagonal Gaussian distribution.} over $\boldsymbol{Z}^{\text{B}}$. Specifically, for document $i$, its encoding process is formulated as:
\begin{equation}
\begin{split}
    \boldsymbol{\mu}^{\text{B}}_i := l^{\text{B}}_{1}(f^{\text{B}}(\boldsymbol{w}^{\text{D}}_i))\\
    \text{diag}(\boldsymbol{\Sigma}^{\text{B}}_i) := l^{\text{B}}_{2}(f^{\text{B}}(\boldsymbol{w}^{\text{D}}_i))\\
    \boldsymbol{z}^{\text{B}}_i \sim {q}^{\text{B}}(\boldsymbol{w}^{\text{D}}_i) := \mathcal{N}\big(\boldsymbol{\mu}^{\text{B}}_i, \text{diag}(\boldsymbol{\Sigma}^{\text{B}}_i)\mathbbm{I}\big)\\
    \text{Output with}~\boldsymbol{z}^{\text{B}}_i:= \text{Softmax}\big(\boldsymbol{z}^{\text{B}}_i\big)
\end{split}~\label{eqn:internal_bow_encoder_doc}
\end{equation}
Here, $f^{\text{B}}: \mathbbm{R}^{ V}\rightarrow\mathbbm{R}^{L}$ is a multi-layer perceptron (MLP) for computing a shared hidden layer output from $\boldsymbol{w}^{\text{D}}_i$, where $L$ is the number of hidden neurons. The functions $l^{\text{B}}_{1}, l^{\text{B}}_{2}: \mathbbm{R}^{L}\rightarrow\mathbbm{R}^{K}$ are two linear layers for respectively predicting the posterior mean and variance vectors: $\boldsymbol{\mu}^{\text{B}}_i, \text{diag}(\boldsymbol{\Sigma}^{\text{B}}_i) \in \mathbbm{R}^{K}$ that generate the document-topical embedding $\boldsymbol{z}^{\text{B}}_i$ for document $i$. The symbol $\mathbbm{I}\in\mathbbm{R}^{K\times K}$ denotes an identity matrix used to create the diagonal Gaussian. Finally, it is common for NTMs to transform the topical embedding $\boldsymbol{z}^{\text{B}}_{i}$ into a probability distribution of topics for each document using Softmax.

As for the NVTM's decoder network, it reconstructs 
$\boldsymbol{W}^{\text{D}}$ using the document-topic matrix $\boldsymbol{Z}^{\text{B}}$ and the word-topic matrix $\boldsymbol{\Phi}^{\text{B}}$ as follows:
\begin{align}
\begin{split}
\boldsymbol{W}^{\text{D}} \sim \text{Multi}\big(\text{Softmax}\big(\boldsymbol{Z}^{\text{B}}{(\boldsymbol{\Phi}^{\text{B}})}^{{T}}\big),\boldsymbol{N}\big)\\
    \end{split}
    \label{eqn:decoder}
\end{align} where 
$\boldsymbol{N}$ is a vector of document lengths and $\boldsymbol{\Phi}^{\text{B}}$, capturing the latent topics of each word in the vocabulary, can also be viewed as the weight matrix of the (output layer) of the decoder network.
\begin{figure}[t]
\centering
\includegraphics[width=\columnwidth]{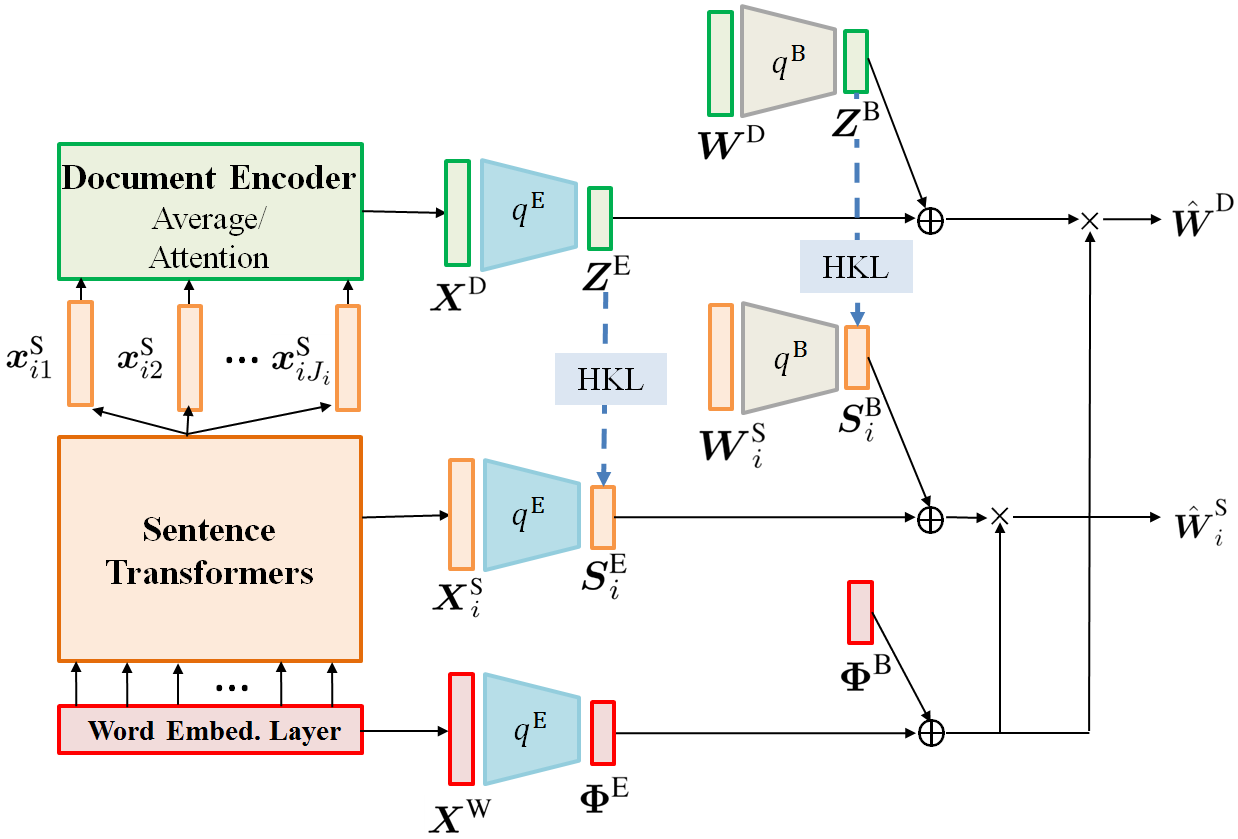}
\caption{Basic Architecture of NAHTM. Each colour highlights a level of information, including the word-level (red), sentence-level (orange) and document-level (green) internal BoW and external semantic information. The blue dash line highlights the hierarchical KL (HKL) constraints over the sentence-document pairs.}
\label{NAHTM_structure}
\end{figure}
\section{Our Model}
Our proposed model, \textbf{N}eural \textbf{A}ttention-aware \textbf{H}ierarchical \textbf{T}opic \textbf{M}odel (NAHTM), comprises (1) two types of encoders, \textit{internal BoW data encoder} and \textit{external knowledge encoder}, that capture latent topics of documents, sentences and words from both internal and external sources, respectively; (2) an attention-aware hierarchical KL divergence that regularizes topical embeddings of sentences with those of their documents. Figure~\ref{NAHTM_structure} shows the basic architecture of NAHTM.

\subsection{Internal BoW Data Encoder}
This encoder, $q^{\text{B}}(\cdot)$, aims to capture the document- and sentence-level BoW data information. The encoding process of the former has been specified in eq.~\eqref{eqn:internal_bow_encoder_doc}, which yields the posterior distributions ${q}^{\text{B}}(\boldsymbol{W}^{\text{D}})$ for the document-topic matrix $\boldsymbol{Z}^{\text{B}}$.

As for encoding the sentence-level BoW data, it is motivated by the fact that sentences convey complete logical statements organized by topics as documents. The difference, as argued by the past research on LDA models, is that sentences are more concise with shorter text and focused topics \cite{balikas-etal-2016-sentlda,balikas-etal-2016-modeling,amoualian-etal-2017-topical}. 
NAHTM encodes the sentence-level BoW data in the same way as it encodes the document-level data except for the final activation function. More specifically, each document is now viewed as a corpus, while each sentence is viewed as a (short) document. For document $i$, its BoW data $\boldsymbol{W}^{\text{S}}_i$, over its sentences, is encoded as ${q}^{\text{B}}(\boldsymbol{W}^{\text{S}}_i)$ with the same encoding process as in eq.~\eqref{eqn:internal_bow_encoder_doc}. Then, the sentence-topical embedding matrix $\boldsymbol{S}^{\text{B}}_i$ for document $i$ is generated as: $\boldsymbol{S}^{\text{B}}_i\sim {q}^{\text{B}}(\boldsymbol{W}^{\text{S}}_i)$.

Based on the well-grounded argument that a sentence should be bound in topics, NAHTM forces $\boldsymbol{S}^{\text{B}}_i$ to be sparse over topics by using Sparsemax \cite{martins-sparsemax-2016,lin-sparsemax-2019} which projects real-valued embeddings into sparse probability vectors: $\boldsymbol{S}^{\text{B}}_i := \text{Sparsemax}(\boldsymbol{S}^{\text{B}}_i)$. Specifically, for the $j^{\text{th}}$ sentence of document $i$, its embedding $\boldsymbol{s}^{\text{B}}_{ij}$ is converted by Sparsemax as:
\begin{equation}
    \begin{split}
        \text{Sparsemax}(\boldsymbol{s}^{\text{B}}_{ij}) := \text{argmin}_{\boldsymbol{c}}\lVert\boldsymbol{c}-\boldsymbol{s}^{\text{B}}_{ij}\rVert^{2}
    \end{split}
\end{equation} where $\boldsymbol{c}$ lies on the ($K-1$)-dimensional probability simplex. In other words, Sparsemax projects $\boldsymbol{s}^{\text{B}}_{ij}$ from the Euclidean space onto the probability simplex.

\subsection{External Knowledge Encoder}
External semantic knowledge, extracted by pre-trained language models (e.g. BERT \cite{devlin2018bert}) from large general corpora, provides rich prior information regarding the contexts and semantic relatedness of instances for each entity (i.e. document, sentence and word) modelled by NTMs. The language models account for ordering patterns of the entities (i.e. sentence and word orderings), which are complementary to the orderless BoW information captured by the NTMs. Incorporating such knowledge into the NTMs can potentially help them better infer sentences' true topics in scenarios which cannot be distinguished by the BoW information. For example, a pair of sentences with the same word counts might have very different topics due to different word orderings. Meanwhile, another pair without any word overlap might still have strongly correlated topics due to a next-sentence or entailment relationship. Hence, NTMs can be guided to better infer topics of documents as well as the entire set of topics underlying the corpus.

Another advantage of external knowledge is that it can potentially alleviate the data sparsity problem under the BoW modelling, especially at the sentence level. Since sentences have much shorter text compared to documents, therefore, their BoW data is also much sparser with significantly fewer word co-occurrences within each sentence and word overlaps in between. 
To make the learning less affected by the sparse data, external knowledge can be leveraged to calibrate it with prior information regarding the sentences.
NAHTM incorporates three levels of external knowledge in the form of the following pre-trained embeddings for words, sentences and documents. 

\textbf{External word embeddings} $\boldsymbol{X}^{\text{W}}$ are output by the embedding layer of the pre-trained transformer model for each word in the vocabulary. Here, $\boldsymbol{X}^{\text{W}}$ are non-contextualized and untrainable embeddings whose dimension is predefined by the pre-trained transformer and thus, not necessarily equal to the number of topics. The reason for using the non-contextualized word embedding is that the word-level external information is expected to be ``injected'' correspondingly into $\boldsymbol{\Phi}^{\text{B}}$, the factorized and non-contextualized word-topical embeddings in our topic model.

\textbf{External sentence embeddings} $\boldsymbol{X}^{\text{S}}_i$ for the sentences of document $i$ are the aggregated results of the outputs from the last transformer encoder (layer) of the pre-trained model. In this case, the inputs to the pre-trained model consist of the word sequences for each sentence. The outputs are the contextualized embeddings of the words in the sequences. In this paper, we use sentence-transformers\footnote{https://github.com/UKPLab/sentence-transformers}, a programmable framework that provides a variety of pre-trained transformer models for computing sentence embeddings. We adopt the default aggregation strategy of sentence-transformers which is to average all the (output) contextualized word embeddings over the sentence.

\textbf{External document embeddings} $\boldsymbol{X}^{\text{D}}$ can be constructed as either the unweighted or the weighted average of all the sentence embeddings for the same document. Specifically, for the latter case, $\boldsymbol{x}^{\text{D}}_i$ can be obtained as follows:
    \begin{align}
    \begin{split}
    y^{\text{S}}_{ij}&:=\sigma_1(\boldsymbol{x}^{\text{S}}_{ij}\boldsymbol{\Theta}_{1}+\boldsymbol{b}_1)\boldsymbol{\theta}_{2}\\
        \boldsymbol{\alpha}^{\text{S}}_i&:=\sigma_2\Big(\boldsymbol{y}^{\text{S}}_i\Big)\\ \boldsymbol{x}^{\text{D}}_i&:=\sum^{J_i}_{j=1}\alpha^{\text{S}}_{ij}\boldsymbol{x}^{\text{S}}_{ij}
    \end{split}
    \label{eqn:sent_att_weight}
    \end{align} where $\boldsymbol{\alpha}^{\text{S}}_i=[\alpha^{\text{S}}_{i1},...,\alpha^{\text{S}}_{iJ_i}]^{\text{T}}$ is the attention (weight) vector over each sentence. Its $j^{\text{th}}$ element $\alpha^{\text{S}}_{ij}$ is the normalized weight of the $j^{\text{th}}$ sentence in representing document $i$, i.e. $\sum^{J_i}_{j=1}\alpha^{\text{S}}_{ij}=1$;  $\boldsymbol{y}^{\text{S}}_i=[y^{\text{S}}_{i1},...,y^{\text{S}}_{iJ_i}]^{\text{T}}$ is the corresponding unnormalized attention vector; $\boldsymbol{\Theta}_1\in\mathbbm{R}^{M\times M}$,  $\boldsymbol{\theta}_2\in\mathbbm{R}^{M}$ and $\boldsymbol{b}_1\in\mathbbm{R}^{M}$ are the attention parameters with $M$ being the (pre-defined) dimension of the pre-trained embedding; $\sigma_1, \sigma_2$ are the activation functions. For $\sigma_1$, we use the Tanh function for all the experiments. For $\sigma_2$, it can be either the conventional Softmax function or the Sparsemax function which induces sparsity over $\boldsymbol{\alpha}^{\text{S}}_i$ such that unimportant sentences tends to have zero weights in representing the document.

\textbf{Mapping} $\boldsymbol{X}^{\text{W}}, \boldsymbol{X}^{\text{D}}, \boldsymbol{X}^{\text{S}}_i$ \textbf{to the topical space} is done subsequently by the external knowledge encoder to align the dimension and semantic meanings of external embeddings with the topical embeddings. More specifically, each level of the external knowledge data $\boldsymbol{X}\in\{\boldsymbol{X}^{\text{W}}, \boldsymbol{X}^{\text{D}}, \boldsymbol{X}^{\text{S}}_i\}$ is encoded into the corresponding posterior distributions $\mathcal{N}\Big(l^{\text{E}}_{1}\big(f^{\text{E}}({\boldsymbol{X}})\big), l^{\text{E}}_{2}\big(f^{\text{E}}({\boldsymbol{X}})\big)\Big)$. Here, all the symbols have the same meanings as those in eq.~\eqref{eqn:internal_bow_encoder_doc} except that they are dedicated to the external knowledge encoder. Correspondingly, we denote the topical embeddings generated by this encoder for the word, sentence and document-level external knowledge respectively as: $\boldsymbol{\Phi}^{\text{E}}\sim q^{\text{E}}(\boldsymbol{X}^{\text{W}}):=\mathcal{N}^{\text{E}}_{\boldsymbol{X}^{\text{W}}}, \boldsymbol{S}^{\text{E}}_i\sim q^{\text{E}}(\boldsymbol{X}^{\text{S}}_i):=\mathcal{N}^{\text{E}}_{\boldsymbol{X}^{\text{S}}_i}$ and $\boldsymbol{Z}^{\text{E}}\sim q^{\text{E}}(\boldsymbol{X}^{\text{D}}):=\mathcal{N}^{\text{E}}_{\boldsymbol{X}^{\text{D}}}$. NAHTM combines the internal and external topical embedding matrices as follows:
\begin{equation}
\begin{split}
\boldsymbol{\Phi}^{\text{Comb}} &:= \boldsymbol{\Phi}^{\text{B}} + \gamma_1\boldsymbol{\Phi}^{\text{E}},\\
\boldsymbol{S}^{\text{Comb}}_i &:= \text{Sparsemax}(\boldsymbol{S}^{\text{B}}_i + \gamma_2\boldsymbol{S}^{\text{E}}_i),\\
\boldsymbol{Z}^{\text{Comb}} &:= \text{Softmax}(\boldsymbol{Z}^{\text{B}} + \gamma_3\boldsymbol{Z}^{\text{E}})
\end{split}\label{eqn:combined_embedding}
\end{equation} where $\gamma_1,\gamma_2$ and $\gamma_3$ are the hyper-parameters that control the influence of external knowledge in calibrating the internal one at the different levels.

\subsection{Attention-Aware KL Divergence}\label{sec:att_kl}
NAHTM makes use of the hierarchical structure of documents by setting the posterior distributions of the document topical embeddings as the priors to their sentences' topical embeddings. More specifically, for document $i$, its dedicated KL divergence term is $Q^{\text{D}}_i:=\beta_{0}\Big(\text{KL}\big[{q}(\boldsymbol{z}^{\text{B}}_i|\boldsymbol{w}^{\text{D}}_i)||{p}(\boldsymbol{z}^{\text{B}}_i)\big]+\text{KL}\big[{q}(\boldsymbol{z}^{\text{E}}_i|\boldsymbol{x}^{\text{D}}_i)||{p}(\boldsymbol{z}^{\text{E}}_i)\big]\Big)$, while for the $j^{\text{th}}$ sentence of document $i$, the dedicated KL term is $Q^{\text{S}}_{ij}:=\beta_{1}\Big(\text{KL}\big[{q}(\boldsymbol{s}^{\text{B}}_{ij}|\boldsymbol{w}^{\text{S}}_{ij})||{q}(\boldsymbol{z}^{\text{B}}_i|\boldsymbol{w}^{\text{D}}_i)\big]
+\text{KL}\big[{q}(\boldsymbol{s}^{\text{E}}_{ij}|\boldsymbol{x}^{\text{S}}_{ij})||{q}(\boldsymbol{z}^{\text{E}}_i|\boldsymbol{x}^{\text{D}}_i)\big]\Big)$. Here, $\beta_0$ and $\beta_1$ are the hyper-parameters that control the regularization strengths of the corresponding KL terms in the ELBO. Note that all the embeddings involved in the above KL terms are \textit{unnormalized}; in other words, they have not been transformed by the Softmax/Sparsemax function.

The assumption behind the above hierarchical structure of KL divergence terms is straightforward: topics of sentences should be somewhat similar to those of their documents. In this case, the degree of the topical similarity constraint enforced into the learning of NAHTM is controlled by $\beta_1$. For example, if $\beta_1$ becomes smaller, the similarity constraint is going to be weakened accordingly.

\textbf{Customising} $\beta_1$ \textbf{with attention weights} is an alternative method we propose that allows for adaptive control on the regularization strengths of the KL divergence terms specific to individual sentence-document pairs. The intent of this method is that the semantic relevance of sentences towards the document, as revealed by the external knowledge, should also be indicative of their topical relevance in the corpus. There are two strategies for implementing this method.

In \textbf{strategy 1}, NAHTM integrates each unnormalized attention weight $\boldsymbol{y}^{\text{S}}_{ij}$, computed from eq.~\eqref{eqn:sent_att_weight}, into the sentence-document KL terms with respect to the pre-trained sentence embedding as:
\begin{equation}
    \begin{split}
        Q^{\text{S}}_{ij}&:=\beta_{1}\sigma({y}^{\text{S}^{'}}_{ij})\Big(\text{KL}\big[{q}(\boldsymbol{s}^{\text{B}}_{ij}|\boldsymbol{w}^{\text{S}}_{ij})||{q}(\boldsymbol{z}^{\text{B}}_i|\boldsymbol{w}^{\text{D}}_i)\big]+\\&
        \text{KL}\big[{q}(\boldsymbol{s}^{\text{E}}_{ij}|\boldsymbol{x}^{\text{S}}_{ij})||{q}(\boldsymbol{z}^{\text{E}}_i|\boldsymbol{x}^{\text{D}}_i)\big]\Big)+
        \lambda_0\lVert{y}^{\text{S}^{'}}_{ij}-{y}^{\text{S}}_{ij}\rVert^2
    \end{split}\label{eqn:custom_beta1_s1}
\end{equation}
where $y^{\text{S}^{'}}_{ij}$ is a latent variable to be learned to control the KL terms specific to the sentence $j$ of document $i$, and is constrained to be close to $y^{\text{S}}_{ij}$; $\lambda_0$ is a hyper-parameter that controls the strength of the constraint; $\sigma({y}^{\text{S}^{'}}_{ij})$ is the corresponding normalized result by either the Softmax or the Sparsemax function. Eq.~\eqref{eqn:custom_beta1_s1} is essentially a ``soft'' version of the strategy that directly uses the normalized attention weight as the controlling parameter, i.e. $\beta_1\alpha^{\text{S}}_{ij}$, for the KL terms across the $J_i$ sentences.

In \textbf{strategy 2}, instead of learning the attention weights $\boldsymbol{y}^{\text{S}}_i$, they are first pre-computed based on the pre-trained sentence and document embeddings by
$y^{\text{S}}_{ij} := -\lVert\boldsymbol{x}^{\text{S}}_{ij}-\boldsymbol{x}^{\text{D}}_i\rVert$
where $\boldsymbol{x}^{\text{D}}_i:=\frac{1}{J_i}\sum^{J_i}_{j=1}\boldsymbol{x}^{\text{S}}_{ij}$. In this case, the unnormalized attention weight $y^{\text{S}}_{ij}$ for sentence $j$ is the negative Euclidean distance between its embedding and the document embedding which is the centroid across all the sentences. The further away the embedding $\boldsymbol{x}^{\text{S}}_{ij}$ is from the centroid $\boldsymbol{x}^{\text{D}}_{i}$, the smaller the weight. Therefore, the control over strengths of the KL constraints is now adaptive only towards the external knowledge. The rest of the KL computation follows exactly eq.~\eqref{eqn:custom_beta1_s1}. For a ``hard'' version of this strategy, we can subsequently normalize the pre-computed $\boldsymbol{y}^{\text{S}}_i$ to obtain $\boldsymbol{\alpha}^{\text{S}}_i$ and then, directly use them as the controlling parameters (same as in strategy 1).

\subsection{Training Objective}
In summary, the training objective of NAHTM is:
\begin{equation}
\small
\begin{split}
 &\mathbbm{E}_{\boldsymbol{Z}^{\text{Comb}}}\big[\text{log}~{p}_{\boldsymbol{\Phi}^{\text{Comb}}}(\boldsymbol{W}^{\text{D}}|\boldsymbol{Z}^{\text{Comb}})\big] + \gamma_4\sum^{I}_i\mathbbm{E}_{\boldsymbol{S}^{\text{Comb}}_i}\big[\text{log}\\&~{p}_{\boldsymbol{\Phi}^{\text{Comb}}}(\boldsymbol{W}^{\text{S}}_i|\boldsymbol{S}^{\text{Comb}}_i)\big]-\sum^{I}_{i}\sum^{J}_{j}Q^{\text{S}}_{ij}-\sum^{I}_{i}Q^{\text{D}}_{i}-\sum^{V}_{v}Q^{\text{W}}_{v}
\end{split}
\end{equation}
where the likelihood ${p}_{\boldsymbol{\Phi}^{\text{Comb}}}$ is modelled by the decoder network as in eq.~\eqref{eqn:decoder} with the weight matrix now being $\boldsymbol{\Phi}^{\text{Comb}}$; $\gamma_4$ controls the influence of the sentence-level training loss; $Q^{\text{D}}_i$ and $Q^{\text{S}}_{ij}$ are respectively the document- and sentence-level KL terms specified in Section~\ref{sec:att_kl}; $Q^{\text{W}}_{v}$ consists of the regularization and KL terms respectively for the internal and external embeddings for each word $v$ from the vocabulary; $Q^{\text{W}}_{v}:=\lambda_1\lVert\boldsymbol{\Phi}^{\text{B}}\rVert+\beta_2\text{KL}[q(\boldsymbol{\phi}^{\text{E}}_{v}|\boldsymbol{x}^{\text{W}}_v)||p(\boldsymbol{\phi}^{\text{E}}_{v})]$. Again, $\lambda_1$ and $\beta_2$ are the controlling hyper-parameters for the respective terms. 
\section{Experimental Setup}
\paragraph{Data}
We evaluate the efficacy of NAHTM using four real-world corpora from a variety of domains, Wikitext-103 \cite{nan-etal-2019-topic}, 20NewsGroup \cite{Srivastava_Sutton_2017}, COVID-19 open research dataset (CORD)\footnote{https:~/~~/~www.kaggle.com~/~allen-institute-for-ai~/~CORD-19-research-challenge} and NIPS\footnote{https:~/~~/~www.kaggle.com~/~benhamner~/~nips-papers} datasets. Table \ref{table:data_summary} summarises the key statistics of these datasets. For the Wikitext-103 dataset, we only use the introduction part of each document, named the WikiIntro dataset, to examine the efficacy of our model on short text. For the CORD dataset, we randomly sampled 20,000 documents from its original corpus for our experiments, which is named CORD20K. 

We adopt the same training-validation-testing split ratios and preprocessing steps from the original papers for the Wikitext-103 (i.e. 70\%-15\%-15\%) and 20NewsGroup (i.e. 48\%-12\%-40\%) datasets. To construct their vocabularies for topic modelling, we follow the same stemming/lemmatization and stopword removing procedures of \citet{hoyle-etal-2020-improving}. 

As for the CORD20K and NIPS datasets, we set the split ratio to be 60\%-20\%-20\%, and use the most frequent 10,000 words (with stemming and stop-words removed) as the vocabulary. Specifically, we apply WordNet lemmatizer and English stopword list (both from the NLTK\footnote{https://www.nltk.org/} toolkit) to preprocess their text. Finally, we use the first 50 sentences of each document from the two datasets for the experiments.

Furthermore, to extract the sentence embedding from the external pre-trained transformer, we have not done any pre-processing (neither stemming/lemmatization nor stopword removing) on the sentence text for all the datasets, as required by the sentence-transformers library. Doing so guarantees that the pre-trained language model can capture the full contextual information within each sentence.

\begin{table}[h]
\centering
\small
\begin{tabular}{cccc}
\hline
&$I$ & $V$ & Avg$N$ \\
\hline
20NewsGroup&17,992&1,995&87.1\\
WikiIntro&28,532&20,000&124.11\\
CORD20K&20,000&10,000&1127.01\\
NIPS&7,241&10,000&1139.26\\
\hline
\end{tabular}
\caption{Summary of the four datasets in terms of the document number $I$, vocabulary size $V$ and average document length Avg$N$ (first 50 sentences).}\label{table:data_summary}
\end{table}

\paragraph{Evaluation Metrics}

We seek to perform two types of evaluation on our model. The first one is the ability to discover a set of latent topics that are meaningful and useful to human. To achieve this, we look at topic coherence with the normalized mutual pointwise information (NPMI)~\cite{aletras2013evaluating,lau2014machine}, which is positively correlated with human judgments on topic quality. Specifically, we calculate the NPMI scores on both the internal corpora from Table \ref{table:data_summary}, and a large external corpus. For each calculation, we first select the top 10 words under each topic based on the (sorted) values in each column of the word embedding matrix $\boldsymbol{\Phi}^{\text{Comb}}$. Then, we calculate and average the internal NPMI scores of each topic as in \cite{bouma2009normalized} with a sliding window of size 10 and the training data as the reference corpus. 
As for the external NPMI, it is calculated using Palmetto\footnote{https://aksw.org/Projects/Palmetto.html} over a large external English Wikipedia dump.

The second metric we use is perplexity, a popular criterion that evaluates how well topic models fit the BoW data, which is calculated over the testing data 
as\footnote{Our perplexity calculation follows that of \cite{pmlr-v48-miao16,pmlr-v70-miao17a,card-etal-2018-neural} but for fair comparisons, the KL divergence is not included for calculating the perplexity as different models weight KL differently.}: $\text{exp}(-\frac{1}{I}\sum^{I}_{i=1}\frac{1}{N_i}\sum^{N_i}_{n=1}{p}(\boldsymbol{w}^{\text{D}}_i))$ where $N_i$ is the length of document $i$, and ${p}(\boldsymbol{w}^{\text{D}}_i)$ is the log-likelihood of the model on the document's BoW data. It is approximated as ${p}(\boldsymbol{w}^{\text{D}}_i)\approx{p}_{\boldsymbol{\Phi}^{\text{Comb}}}(\boldsymbol{w}^{\text{D}}_i|\boldsymbol{z}^{\text{Comb}}_i)$. Here, $\boldsymbol{\Phi}^{\text{Comb}}$ combines the decoder weights $\boldsymbol{\Phi}^{\text{B}}$ and the posterior mean for $\boldsymbol{\Phi}^{\text{E}}$, while $\boldsymbol{z}^{\text{Comb}}_i$ combines the posterior means for $\boldsymbol{z}^{\text{B}}_i$ and $\boldsymbol{z}^{\text{E}}_i$, as shown in eq.~\eqref{eqn:combined_embedding}, at the first training step after the model has converged with respect to the validation perplexity.

\paragraph{Baselines}
We compare NAHTM with various state-of-the-art baselines which can be broadly categorised into 1) BoW-based autoencoder models without external knowledge, which include Scholar (without meta-data) \cite{card-etal-2018-neural}, W-LDA \cite{nan-etal-2019-topic}, GSM \cite{pmlr-v70-miao17a}, ETM \cite{dieng-etal-2020-topic} and RRT \cite{tian-etal-2020-learning}; 2) models with external knowledge learned by pre-trained language models including CTM \cite{bianchi-etal-2021-pre} and BAT \cite{hoyle-etal-2020-improving}. GSM is similar to Scholar but with a simpler encoder-decoder structure. ETM further factorizes the topic-word distribution matrix into multiplication of topic and word embedding vectors. RRT proposes a new reparameterization trick for Dirichlet distributions over the document embeddings.

As for the implementations and settings of the baselines, we use their official codes and default settings obtained from their official Github repositories, except for GSM whose original code is unavailable. In this case, we re-implement GSM by referring to other credible sources\footnote{https://github.com/zll17/Neural\_Topic\_Models}. For BAT, we use its enhanced versions with Scholar (i.e. BAT+Scholar) and W-LDA (i.e. BAT+W-LDA) from its implementation. To allow for fair comparisons, we tune the major parameters for all the models, including the numbers of hidden layers \{1, 2, 3\} and hidden neurons \{100, 300, 600\}, the learning rate \{0.001, 0.002, 0.005, 0.01\} and the batch size \{8, 20, 200\}. The ranges of the above hyper-parameters are the most common ones set by the baselines in their own implementations. We generally found that 1 hidden layer with 300 neurons, a learning rate of 0.002 and a batch size of 20 yields the best overall perplexity and NPMI performance across the models. For CTM and BAT that incorporate external knowledge as NAHTM does, they use the same pre-trained transformers as NAHTM, including ``bert-base-uncased'', ``distilbert-base-uncased'' and ``roberta-base'' from the Huggingface Transformers models\footnote{https://huggingface.co/transformers/pretrained\_models.html}.

\subsection{NAHTM Settings}

The hyper-parameters of NAHTM control the influence of its different components and we use the validation dataset to optimize their values in terms of the validation perplexity. We find the same values generally hold, across the datasets, for $\gamma_1=0.01$ and $\gamma_2=0.001$ that control the effects of external embeddings for words and documents respectively. On the other hand, $\gamma_3$ and $\gamma_4$ were respectively tuned over the sets of candidate parameter values $\{0.001, 0.01, 0.1, 1\}$ and $\{0.01, 0.1, 1, 5\}$ to different optimal values for different datasets; $\beta_0$, $\beta_1$ and $\beta_2$ were all optimized over the value set $\{0.001, 0.01, 0.1, 1\}$, which, compared with the KL annealing approach~\cite{bowman-etal-2016-generating}, is much more efficient albeit sub-optimal; $\lambda_0$ and $\lambda_1$ were both tuned over the value set  $\{0.01, 0.1, 1, 5, 10\}$. To allow for fair comparisons with the baselines, we set the number of hidden layers for both the internal BoW and the external knowledge encoders to be 1, the number of neurons to be 300, the batch size to be 20, and the learning rate to be 0.002 for all the experiments. 

\begin{table}[t]
\scriptsize
\centering
\begingroup
\setlength{\tabcolsep}{2pt} 
\renewcommand{\arraystretch}{1} 
  \begin{tabular}{lc|c|c|c}
    \toprule
     Models& WikiIntro & 20News & NIPS & CORD20K\\
    \midrule
    &\multicolumn{3}{c}{\textbf{External NPMI}}&\\
Scholar &  0.116~/~0.109 & 0.028~/~0.009 & 0.013~/~-0.017&0.036~/~0.007\\
W-LDA &0.108~/~0.096  &0.034~/~0.012 & 0.013~/~-0.022&0.031~/~-0.001\\
GSM &0.111~/~0.094  &0.018~/~-0.006 & -0.022~/~-0.062&0.023~/~-0.011\\
    ETM & 0.082~/~0.064 & 0.003~/~-0.014 & -0.062~/~-0.097&0.008~/~-0.032\\
RRT &0.118~/~0.094& 0.009~/~-0.010 & 0.011~/~-0.031&0.027~/~-0.013\\
BAT+Scholar &0.122~/~0.113  &0.039~/~0.017 & 0.014~/~-0.011&0.041~/~0.008\\
BAT+W-LDA &0.119~/~0.108  &\underline{0.041}~/~0.017 & 0.014~/~-0.010&0.039~/~0.004\\
   CTM & 0.125~/~0.110 & 0.040~/~\underline{0.018} & 0.013~/~-0.013&0.036~/~0.010\\
    \hhline{-----}
    $\text{NAHTM}_{\text{KL}}$&0.132~/~0.114 & 0.035~/~0.013 &0.014~/~-0.006&0.039~/~0.008\\
     $\text{NAHTM}_{\text{HKL}}$&0.138~/~0.119 & 0.037~/~0.015 &0.014~/~-0.005&0.042~/~0.010\\
     $\text{NAHTM}_{\text{S1}}$ &\underline{0.145}~/~\underline{0.132} & 0.040~/~0.018 & \underline{0.015}~/~\underline{-0.003}&\underline{0.044}~/~\underline{0.011}\\
     $\text{NAHTM}_{\text{S2}}$ &\textbf{0.149}~/~\textbf{0.135} & \textbf{0.042}~/~\textbf{0.020} & ~\textbf{0.016}~/~\textbf{0.007}&\textbf{0.046}~/~\textbf{0.015}\\

        \midrule
   &\multicolumn{3}{c}{\textbf{Internal NPMI}}&\\
Scholar &  0.427~/~0.411 & 0.256~/~0.224 & 0.285~/~0.256&0.349~/~0.294\\
W-LDA &0.422~/~0.409  &0.249~/~0.220&0.294~/~0.260&0.337~/~0.286\\
GSM &0.411~/~0.398  &0.225~/~0.198 &0.289~/~0.242&0.318~/~0.239\\
    ETM & 0.393~/~0.384 & 0.190~/~0.165 &0.222~/~0.176&0.247~/~0.188\\
RRT &0.427~/~0.414& 0.254~/~0.218 &0.249~/~0.212&0.343~/~0.265\\
BAT+Scholar &0.439~/~0.421  &0.268~/~0.243 & 0.298~/~0.260&0.361~/~0.303\\
BAT+W-LDA &0.432~/~0.416  &0.264~/~0.240 & 0.309~/~0.263&0.353~/~0.294\\
    CTM & 0.438~/~\underline{0.422} & 0.269~/~0.246 &0.295~/~0.254&0.356~/~0.290\\
    \hhline{-----}
    $\text{NAHTM}_{\text{KL}}$&0.437~/~0.416 & 0.263~/~0.245 &0.302~/~0.262&0.356~/~0.296\\
     $\text{NAHTM}_{\text{HKL}}$ &0.441~/~0.418 & 0.266~/~0.249 &0.306~/~0.264&0.358~/~0.301\\
     $\text{NAHTM}_{\text{S1}}$ &\underline{0.442}~/~0.421 & \underline{0.273}~/~\underline{0.251} & \underline{0.311}~/~\underline{0.267}&\underline{0.363}~/~\underline{0.307}\\
     $\text{NAHTM}_{\text{S2}}$ &\textbf{0.446}~/~\textbf{0.425} & \textbf{0.279}~/~\textbf{0.256} & \textbf{0.319}~/~\textbf{0.272}&\textbf{0.370}~/~\textbf{0.315}\\
        \midrule
        &\multicolumn{3}{c}{\textbf{Perplexity}}&\\
        Scholar &  2,071~/~1,938 & 978~/~936 &1,686~/~1,584&1,562~/~1,489\\
W-LDA &2,243~/~2,091  &951~/~927 &1,754~/~1,632&1,580~/~1,517\\
    GSM &2,465~/~2,176  &1,114~/~1,032 &1,796~/~1,726&1,611~/~1,565\\
    ETM & 2,184~/~2,106 & 1.045~/~985 &1,722~/~1,639&1,559~/~1,523\\
RRT &2,274~/~2,123& 984~/~955 &1,748~/~1,677&1,625~/~1,570\\
BAT+Scholar &1,963~/~1,872  & 947~/~924 &\textbf{1,642}~/~\textbf{1,534}&\underline{1,496}~/~\textbf{1,423}\\
BAT+W-LDA &2,014~/~1,934  & 936~/~\underline{903} &1,715~/~1,603&1,528~/~1,476\\
CTM & 1,924~/~1,860 & 972~/~933 &1,726~/~1,641&1,568~/~1,519\\
    \hhline{-----}
    $\text{NAHTM}_{\text{KL}}$ &1,987~/~1,898 & 959~/~941& 1,752~/~1,670&1,552~/~1,502\\
     $\text{NAHTM}_{\text{HKL}}$ &1,949~/~1,882 & 939~/~916& 1,737~/~1,653&1,534~/~1,483\\
     $\text{NAHTM}_{\text{S1}}$ &\underline{1,911}~/~\underline{1,854} & \underline{921}~/~907& 1,704~/~1,634 &1,512~/~1,469\\
     $\text{NAHTM}_{\text{S2}}$ &\textbf{1,878}~/~\textbf{1,825} & \textbf{905}~/~\textbf{881}& \underline{1,652}~/~\underline{1,567}&\textbf{1,491}~/~\underline{1,450}\\
        \bottomrule
  \end{tabular}
  \endgroup
  \caption{Results of average external and internal NPMI, and average test perplexity for each model with 50 and 200 topics, whose results are respectively on the left and right side of ``/'' in the table. Lower perplexity and higher NPMI scores mean better performance. Bold and underlined values indicate the best and second performing models in each dataset/metric setting.}
  \label{tab:metric_results}
\end{table}

\begin{table*}[t]
\small
\centering
\begingroup
\setlength{\tabcolsep}{2pt} 
\renewcommand{\arraystretch}{1} 
  \begin{tabular}{ll|l|l}
    \toprule
     &Topics&Models& Topic Words\\
    \midrule
& \multirow{2}{0.7in}{Optimization} &BAT+Scholar&\textit{problem}, siam, np, solve, optimization, loss, min, \textit{consider}, convex, hard\\
\hhline{~~~-}
&&$\text{NAHTM}_{\text{S2}}$&loss, gradient, minimization, optimization, solution, min, convex, np, descent, objective\\
\hhline{~---}
NIPS&Neural&BAT+Scholar&layer, neural, deep, convolutional, network, gradient, architecture, \textit{visual}, \textit{input}, recurrent\\
\hhline{~~~-}
&Networks&$\text{NAHTM}_{\text{S2}}$&network, neural, layer, deep, backpropagation, recurrent, convolutional, descent,\\
&&&feedforward, gradient\\
            \midrule
&&BAT+Scholar&infection, covid, \textit{disease}, virus, coronavirus, epidemic, h7n9, vaccine, \textit{patient}, \textit{cases}\\
\hhline{~~~-}
&COVID&$\text{NAHTM}_{\text{S2}}$&\textit{disease}, covid, virus, infected, coronavirus, vaccine, influenza, \textit{patient}, epidemic,\\
CORD&&&respiratory\\
\hhline{~---}
20K&&BAT+Scholar&quarantine, \textit{school}, lockdown, \textit{unemployed}, \textit{study}, control, province, \textit{holiday}, social, city\\
\hhline{~~~-}
&Quarantine&$\text{NAHTM}_{\text{S2}}$&quarantine, lockdown, curfew, \textit{unemployment}, fatality, distancing, \textit{holiday}, social,\\
&&&province, \textit{inequity}\\
 \bottomrule
  \end{tabular}
  \endgroup
  \caption{Top 10 topic words extracted by NAHTM and BAT+Scholar from the NIPS and CORD20K datasets where the italic words are those either too common or less relevant (compared to the other words) to the topics.}\label{tab:topic_examples}
  \end{table*}

\begin{table}[t]
\scriptsize
\scalebox{1.15}{
\centering
\begingroup
\setlength{\tabcolsep}{2pt} 
\renewcommand{\arraystretch}{1} 
  \begin{tabular}{lc|c|c|c}
    \toprule
     Models& WikiIntro & 20News & NIPS & CORD20K\\
    \midrule
    Scholar&965~(36)&648~(54)&861~(45)&3,157~(76)\\
    W-LDA&940~(42)&619~(70)&896~(40)&3,718~(88)\\
    GSM&1,029~(58)&802~(42)&1,055~(66)&3,348~(102)\\
    ETM&944~(19)&710~(22)&965~(32)&2,952~(57)\\
    RRT&981~(74)&747~(95)&1,271~(128)&3,623~(114)\\
    BAT+Scholar&934~(45)&\underline{566}~(63)&\underline{828}~(54)&\textbf{2,256}~(68)\\
    BAT+W-LDA&\underline{915}~(33)&\textbf{548}~(56)&874~(61)&2,744~(93)\\
    CTM&952~(82)&1,025~(51)&1,484~(110)&3,298~(121)\\
    $\text{NAHTM}_{\text{S2}}$&\textbf{892}~(78)&605~(84)&\textbf{788}~(94)&\underline{2,469}~(146)\\
            \bottomrule
  \end{tabular}
  \endgroup
  }
  \caption{With 50 topics, the sentence-level test perplexity results (with standard deviations shown inside brackets) of each model on the BoW data of each of the top 25 sentences within every document.}\label{tab:metric_results1}
  \end{table}

\section{Results and Discussion}

Following the settings from the previous section, we proceed to conduct both quantitative and qualitative experiments to evaluate NAHTM. For the quantitative experiments, we report the results of the average external and internal NPMI, and the average test perplexity over 5 runs with different random seeds for initialization. Moreover,  within each run, all the models were learned twice with 50 and 200 topics each time, and their corresponding metric scores have been summarized in Table \ref{tab:metric_results}. 

It can be observed that NAHTM, equipped with either the $\beta_1$-customising strategy 1 or 2 from Section \ref{sec:att_kl} (i.e. $\text{NAHTM}_{\text{S1}}$ and $\text{NAHTM}_{\text{S2}}$), and with uncustomised hierarchical KL (i.e. $\text{NAHTM}_{\text{HKL}}$), is overall more coherent, in terms of both types of NPMI, than the baseline models and NAHTM with independent KL terms for sentences and documents (i.e. $\text{NAHTM}_{\text{KL}}$). Especially when comparing with BAT+\{Scholar,W-LDA\} and CTM, which also have leveraged external pre-trained knowledge, NAHTM manages to achieve not only higher topic coherence but also lower perplexity on the test data in general. The only exception is on the NIPS dataset where NAHTM follows BAT+Scholar closely on the perplexity.

In addition to its efficacy on the document-level BoW reconstruction, NAHTM is also able to achieve state-of-the-art reconstruction performance at the sentence level. We illustrate this by treating each sentence (from the beginning 50 sentences) as a short document, and applying all the NTMs models to reconstruct their BoW data. In this case, we focus on the test perplexity of the different models on the sentences, reporting the sentence-level perplexity from NAHTM with its inference jointly performed over the document- and sentence-level log-likelihood under the hierarchical KL constraint). Table \ref{tab:metric_results1} shows that at 50 topics, the best variant $\text{NAHTM}_{\text{S2}}$, with respect to the document-level BoW reconstruction, remains competitive on the sentence-level reconstruction task. This suggests that NAHTM, with its attention-aware hierarchical KL regularization, can effectively infer both the document- and sentence-level neural topic models it contains, and enables the latter model to be robust against the sparse sentence-level BoW data.

Furthermore, we conduct an ablation study on the importance of different external knowledge components in contributing to the robustness of NAHTM when dealing with the sparse sentence data. We find from Table \ref{tab:ablation_study} that the external word embeddings are the most important components for enhancing the performance of NAHTM, while the external sentence embeddings are the least important. Despite this, we can still see that the performance of NAHTM is significantly influenced by all the three types of external knowledge.

Finally, to gain a more intuitive view on how well NAHTM has discovered the underlying topics, we show in Table \ref{tab:topic_examples} the top 10 words under each of the four example topics extracted by NAHTM from the NIPS and CORD20K datasets. These four topics are Optimization and Neural Networks from the NIPS dataset, and COVID and Quarantine from the CORD20K dataset. It can be observed that the topics discovered by NAHTM tend to be more coherent and less likely to contain common and irrelevant words which can be found from the topic word lists extracted by BAT+Scholar. 
\begin{table}[t]
\scriptsize
\scalebox{1.10}{
\centering
\begingroup
\setlength{\tabcolsep}{2pt} 
\renewcommand{\arraystretch}{1} 
  \begin{tabular}{lc|c|c|c}
    \toprule
     Models& WikiIntro & 20News & NIPS & CORD20K\\
    \midrule
    $\text{NAHTM}_{\text{S2}}$&892~(78)&605~(84)&788~(94)&2,469~(146)\\
     $\text{NAHTM}_{\text{S2}}$-$\boldsymbol{X}^{\text{W}}$&1,173~(142)&981~(119)&1,035~(182)&3,238~(165)\\
     $\text{NAHTM}_{\text{S2}}$-$\boldsymbol{X}^{\text{D}}$&976~(67)&711~(79)&890~(86)&2,894~(111)\\
     $\text{NAHTM}_{\text{S2}}$-$\boldsymbol{X}^{\text{S}}$&950~(42)&683~(60)&848~(62)&2,726~(98)\\
    \bottomrule
  \end{tabular}
  \endgroup
  }
  \caption{With 50 topics, an ablation study on the effects of removing different external knowledge components on the test perplexity performance of NAHTM.}\label{tab:ablation_study}
  \end{table}
  
\section{Conclusion}
In this paper, we have proposed NAHTM, a VAE-based neural topic model with attention-aware hierarchical KL divergence imposed on the pairs of documents and sentences. NAHTM incorporates both the internal BoW data information and the external pre-trained knowledge for refining the topical embeddings of words, sentences and documents. Both quantitative and qualitative experiments have shown the effectiveness of NAHTM on 1) recovering the BoW data at different levels of granularity and 2) discovering coherent topics, by making use of the hierarchical KL constraints on the sentence-document pairs and the external knowledge. As for the future work, we would like to investigate the possibility of combining NAHTM with neural language models for topic-aware language understanding and content generation.

\section*{Acknowledgements}
Yuan Jin and Wray Buntine were supported by the Australian Research Council under awards DE170100037. Wray Buntine was also sponsored by DARPA under agreement number FA8750-19-2-0501.

\bibliography{anthology}
\bibliographystyle{acl_natbib}

\end{document}